%% file: egpaper_final.tex
\documentclass[10pt,twocolumn,letterpaper]{article}

\usepackage{iccv}
\usepackage{times}
\usepackage{epsfig}
\usepackage{graphicx}
\usepackage{amsmath}
\usepackage{amssymb}

\usepackage{graphicx}
\usepackage{amsmath}
\usepackage{amssymb}
\usepackage{booktabs}
\usepackage[dvipsnames]{xcolor}

\usepackage[pagebackref=true,breaklinks=true,letterpaper=true,colorlinks,bookmarks=false]{hyperref}

\iccvfinalcopy

\usepackage[capitalize]{cleveref}
\crefname{section}{Sec.}{Secs.}
\Crefname{section}{Section}{Sections}
\Crefname{table}{Table}{Tables}
\crefname{table}{Tab.}{Tabs.}

\def\iccvPaperID{10212} 

\usepackage[pagebackref=true,breaklinks=true,letterpaper=true,colorlinks,bookmarks=false]{hyperref}

\usepackage[capitalize]{cleveref}
\crefname{section}{Sec.}{Secs.}
\Crefname{section}{Section}{Sections}
\Crefname{table}{Table}{Tables}
\crefname{table}{Tab.}{Tabs.}

\ificcvfinal\fi

\begin{document}
\definecolor{cl5}{rgb}{1,0.73,0}
\definecolor{cl6}{rgb}{0,0.63,1}
\definecolor{cl2}{rgb}{0.4,0,1.0}
\definecolor{cl3}{rgb}{0,0.8,1}
\definecolor{cl4}{rgb}{0.67,0,0}
\definecolor{cl1}{rgb}{0.93,0.64, 0.06}
\definecolor{cl7}{rgb}{1,0.83, 0}
\definecolor{cl8}{rgb}{0.1,0.48, 0.11}
\definecolor{cl9}{rgb}{0.65,0.15, 0.15}
\definecolor{cl10}{rgb}{0.5,0.5, 1.0}
\definecolor{cl11}{rgb}{0.5,0.5,0.5}

\title{ENTL: Embodied Navigation Trajectory Learner}

\author{
Klemen Kotar $^1$\\
Stanford University\\
\and
Aaron Walsman\\
University of Washington\\
\and
Roozbeh Mottaghi\\
Meta AI\\
}

\maketitle
\ificcvfinal\thispagestyle{empty}\fi


\input{sections/abstract}
\input{sections/01_intro}

\input{sections/02_related}

\input{sections/03_method}

\input{sections/04_experiments}

\input{sections/05_conclusion}

{\small
\bibliographystyle{ieee_fullname}
\bibliography{egbib}
}

\appendix

\input{sections/06_supp}

\end{document}

%% file: sections/abstract.tex
\begin{abstract}
{\textcolor{cl4}{Note: During the process of preparing and optimizing our code for the final release, we observed a high degree of sensitivity to the dataset order and training regime. As a result, we are unable to fully support the claims made in the paper. Therefore, we have decided to withdraw this paper.}} We propose Embodied Navigation Trajectory Learner (ENTL), a method for extracting long sequence representations for embodied navigation. Our approach unifies world modeling, localization and imitation learning into a single sequence prediction task. We train our model using vector-quantized predictions of future states conditioned on current states and actions. ENTL's generic architecture enables the sharing of the the spatio-temporal sequence encoder for multiple challenging embodied tasks. We achieve competitive performance on navigation tasks using significantly less data than strong baselines while performing auxiliary tasks such as localization and future frame prediction (a proxy for world modeling). A key property of our approach is that the model is pre-trained without any explicit reward signal, which makes the resulting model generalizable to multiple tasks and environments. 

\footnotetext[1]{Work mainly done while author was at Allen Institute for AI}
\end{abstract}

%% file: sections/01_intro.tex
\section{Introduction}
Representation learning has shown great success in many AI domains. The approach, common in the NLP field for quite some time, has seen an explosion of popularity through the adaptation of the transformer architecture by work such as BERT~\cite{Devlin2019BERTPO} and GPT~\cite{gpt-3}. This initial success has triggered the creation of models with ever more parameters trained on ever larger datasets. Surprisingly, the growing models not only kept improving with the increased parameter counts and datasets, but even started exhibiting emergent behaviors \cite{FoundtaionModels}. Furthermore, the general recipe of converting data into sequences of tokens and predicting the next token has proven surprisingly universal. Subsequent work such as ViT \cite{vit} and MAE \cite{mae} have adapted this approach to the computer vision domain with great success, and transformer models trained on large scale data have become the de facto computer vision backbone.

\begin{figure}[tp]
    \centering
    \includegraphics[width=9cm]{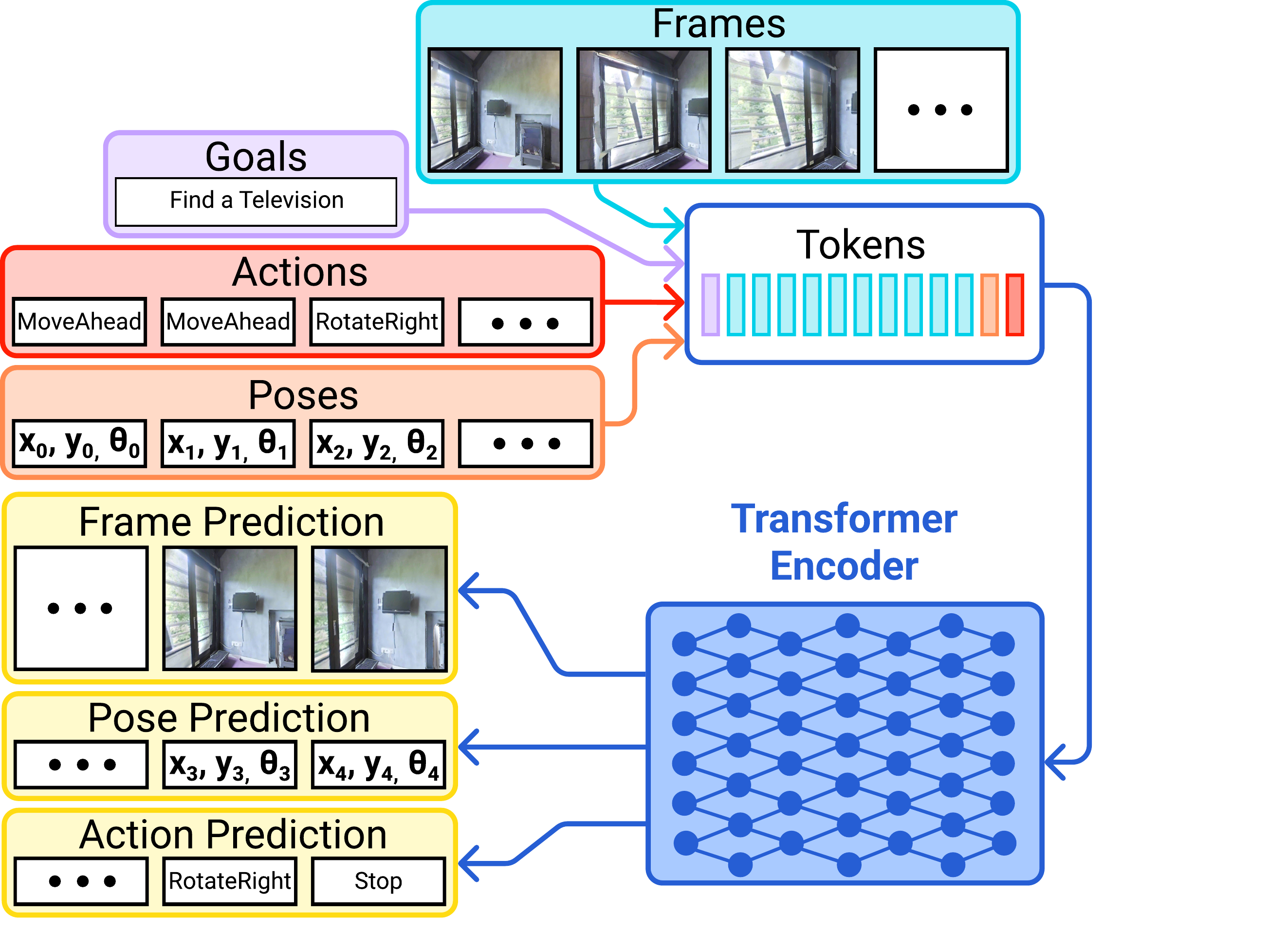}
    \caption{We introduce ENTL a method for extracting long sequence representations for embodied navigation. The proposed architecture enables sharing a spatio-temporal transformer-based backbone across multiple tasks: navigation, localization, and future frame prediction. }
    \label{fig:teaser}
\end{figure} 

Many Embodied AI (EAI) tasks have long horizons and sparse reward signals. This means that EAI models must aggregate information across many time steps, and learn how to attribute the successful or unsuccessful outcomes of entire sequences to individual actions. Since the space and time complexity of transformers grows quadratically with sequence length it has been difficult to apply them to these long sequence tasks. Furthermore, it can be difficult to learn which part of a frame will be significant to a decision the model will have to make many steps down the line, making it hard to decide which information to preserve.

Modeling the environment of the embodied agent is a natural self supervised pre-training task. Various abstractions of this problem, such as the prediction of top down maps \cite{Zhu2022NavigatingTO} or graph-based representations \cite{Gadre2022ContinuousSR} have been studied, but none have yielded a strong universal abstraction. More recently, methods such as \cite{Ren2022LookOT} have explored frame prediction in the RGB space conditioned on camera pose using a buffer of a few frames. These produce visually compelling predictions, but fail to capture the movement dynamics of an embodied agent in the scene by conditioning their models with camera pose instead of agent actions. They also rely on a limited bank of previous frames which only allow the model to extrapolate the textures and shapes of the current view of the room.

We use future frame prediction as a universal navigation pre-training task by conditioning it on agent actions. This encodes the navigability of different areas of the environment within the prediction model, and allows us to use agent walkthroughs of the environment as our sole source of supervision, without the requirement of a known camera pose which might not be available in real world settings. Secondly we utilize an architecture that attends over long sequences (up to 50 steps), allowing the agent to synthesize novel views of the environment from many past examples, and obtaining a richer representation of the environment. Finally, we jointly model future frame prediction, agent pose prediction and action prediction, effectively merging imitation learning and representation learning. This approach directly supervises the agent on how to perform a navigation task, and removes the need for a reward signal.

To ensure that crucial information is not forgotten we propose the prediction of future frames in the original RGB space, avoiding any inductive biases that might come from modeling the embeddings of observations. Since predicting the outputs in RGB pixel space can be unstable and difficult, we utilize the image tokens produced by VQ-GAN~\cite{vqgan}  as our prediction targets. Modern tokenizers such as VQ-GAN can perform near-lossless encodings and reconstructions of RGB images, allowing us to effectively treat them as a reversible transform. 

To overcome the long sequence lengths, we utilize a model architecture that attends across space and time in alternating layers similar to \cite{Bertasius2021IsSA}, allowing the full information of a trajectory to be aggregated without using intractable amounts of memory and computation. We combine this spatio-temporal sequence representation with three separate task heads for localization, future frame prediction and action prediction to maximize parameter sharing and utilize a causal mask to prevent the leakage of information from future states, while still satisfying the input requirements of each head. This design enables us to have a sequence representation that is suitable for all tasks (Figure~\ref{fig:teaser}). 

We apply the self-supervised offline data training recipe and demonstrate that it can produce a model that performs well on embodied navigation tasks. Our model performs competitively against strong Reinforcement Learning (RL) and Imitation Learning (IL) baselines trained on significantly more data on the PointNav~\cite{Anderson2018OnEO} and ObjectNav\cite{batra2020objectnav} tasks. Furthermore, our model performs very well at localization, accruing an average error of only 0.43m after a sequence of 50 steps, using RGB frames only and no motion sensors. Finally, our model produces high-quality, realistic future frame predictions, even in scenes with appearance characteristics well outside its training distribution.

In summary, our key contributions are 1) the formulation of a long sequence future frame prediction problem as a self-supervised task for embodied navigation 2) the proposal of a tokenization scheme and model architecture, which allow us to implement this pre-training 3) the contrastive analysis of several adjacent methods and the necessary pieces required to make our approach work.

%% file: sections/02_related.tex
\section{Related Work}

\textbf{Embodied Representations.} Various works in the literature have addressed representation learning for/from embodied settings. \cite{Pinto2016TheCR} use physical interactions with objects to learn visual representations. \cite{Pathak2017CuriosityDrivenEB} propose a ``curiosity"-based approach to learn representations. \cite{Du2021CuriousRL} also propose a curiosity-driven approach that jointly learns the visual representations and the policy. \cite{Sax2019LearningTN} leverage mid-level representations to improve sample efficiency and generalization for embodied tasks. \cite{HandS} learn representations from agents that play an adversarial game. \cite{Radosavovic2022,Xiao2022} use ego-centric data for visual representation learning, where their goal is to improve the performance of embodied tasks. The focus of these approaches is on \emph{visual} representation learning while our goal is to learn sequences. \cite{Arnab2021ViViTAV,Wu2022MeMViTMM,ViS4mer_ECCV22} encode long videos using transformer models. In contrast, we propose an architecture for embodied settings which serves as a backbone for multiple navigation tasks while handling different input constraints of the tasks. 

Recent work such as \cite{Bonatti2022PACTPC,janner2021offline,chen2021decision} has used transformer-based architectures for interactive settings. We explore applying a vanilla Decision Transformer to our problem and find that predicting a single embedding for future observations is not sufficient. In addition, our approach does not require a reward signal. \cite{gato} focus on performing multiple embodied and language-based tasks jointly. In contrast, we propose an architecture to handle long sequences of observation and actions and we also focus on sample efficiency of training, without jointly modeling language. \cite{Bonatti2022PACTPC} learn representations from robot data in a self-supervised fashion. They do not show results on long-horizon tasks such as navigation. Alternate spatial and temporal layers, vector-quantization of the output space, and handling visually complex environments differentiates our work from \cite{janner2021offline,chen2021decision}.

\textbf{World/Environment Modeling.} \cite{ren2022look,Koh2021PathdreamerAW} learn environment models and generate future frames. Their work is conceptually similar to our future frame prediction. \cite{Guo2020} learn representations for multi-task RL by predicting the future latent embeddings of the observation. \cite{parisi_icml22} show visual representations can be better than state representations for training control policies. In contrast to these works, our aim is to learn representations for sequences of actions, observations, and agent poses. World modeling methods such as Dreamer \cite{Hafner2020MasteringAW} have explored jointly learning a world model and control policy for an interactive agent. We show our method outperforms this approach. Our approach also learns directly from expert trajectories, and does not require a reward signal. Works such as \cite{Ren2022LookOT} predict future frames in RGB space with high success, but only rely on a few buffer frames and are therefore not capable of modeling an entire environment simultaneously or keeping a memory of previously explored locations, thus they are not well suited as a training objective for embodied sequence learning. 
\cite{Suglia2021EmbodiedBA} learn to predict object detections at future states in embodied environments. In contrast we add no inductive bias to our predictor and model the RGB frames of our environment, eliminating the need for ground truth object labels which might not always be available or reliable. \cite{Fang2019SceneMT} use a transformer to embed observations of a scene formed during exploration and access them when making policy decisions. In contrast we train the model to predict future frames based off past observations, forcing the model to synthesize new views and learn the spatial structure of environments. \cite{ramakrishnan2021environment} model embodied navigation trajectories using a transformer. They introduce the inductive bias of splitting their trajectory into zones and predict what zone they are in. Our method simply models action conditioned future frames without any assumptions about the structure of the environment.

\textbf{Representations for Navigation} There are several works addressing representation learning specifically for navigation tasks. For example, \cite{wijmans2019dd} learn visual representations from 2.5 billions of interactions in a point-goal navigation setting. \cite{Ye2020AuxiliaryTS} improve sample efficiency of navigation models by combining representations learned from auxiliary tasks. \cite{embodied-clip} use CLIP representations and show large improvements across a set of embodied tasks. In contrast, we have a generic backbone for multiple tasks.

%% file: sections/03_method.tex
\begin{figure*}[tp]
    \centering
    \includegraphics[width=40pc]{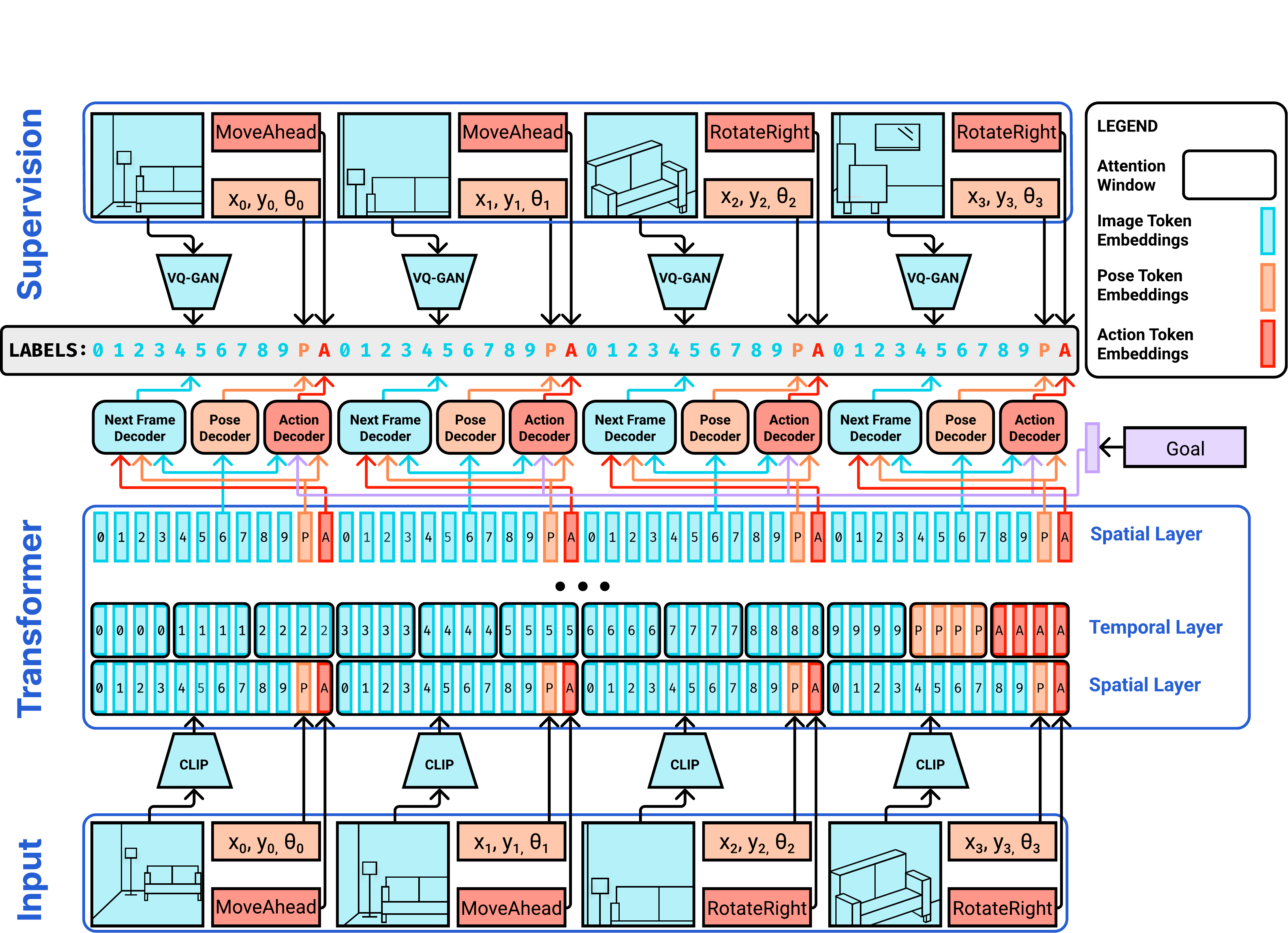}
    \caption{\textbf{ENTL Model Architecture.} Sequences of frames, poses and actions are passed through a spatio-temporal transformer encoder, which aggregates information across space and time in alternating layers to model an entire sequence. Poses and actions are discretized and passed through a learned embedding, while frames are converted to CLIP embeddings. Spatial layers pay attention to all the tokens in a frame, while temporal layers pay attention to the n-${th}$ token of every frame across an entire sequence. In addition to this, a full diagonal causal mask is applied to the attention matrix. After that the representations are passed through three separate transformer heads, which produce the next frame prediction, the current pose of the agent, and the action the agent will take next. The frame self-supervised labels are provided in the form of vector-quantized tokens, while the labels for the actions and pose are just the discretized ground truth values.}
    \label{fig:model}
\end{figure*}

\section{Trajectory Learning}
\textbf{Task definition.}  Our goal in this paper is to learn representations for navigation trajectories (i.e., sequence of images, actions, and poses) of embodied agents. Given a set of trajectories $\mathcal{T}$ of agents navigating in environments, we train a spatio-temportal transformer model to encode these sequences. Each trajectory $\tau \in \mathcal{T}$ is a sequence $(I_i, p_i, a_i)_{i=0}^N$, where $I$, $p$, and $a$ represent image, agent pose, and action, respectively and are collectively called a state. We represent the pose $p$ as $(x,y,\theta)$, the position and orientation of the agent. The objective is to predict the future frame, the action, and the pose of the agent (in a vector-quantized token space described in Section~\ref{sec:token}). 

The challenges are: (1) how to encode long sequences of embodied tasks, (2) how to represent observations, actions, and poses so they can be processed by a single transformer model, (3) and how to train these models without any explicit reward. 

Our architecture enables sharing the backbone across four tasks: future frame prediction, agent localization, object-goal navigation (ObjectNav)~\cite{batra2020objectnav}, and point-goal navigation (PointNav)~\cite{Anderson2018OnEO}. Our approach relies on treating all of the observations our agent collects and the actions it takes as part of the same long contiguous sequence. Deciding which action our agent should take next is achieved by predicting the next action token in the sequence, and localization is achieved by predicting the pose tokens. Future frame prediction can be achieved by appending the desired action token to our current sequence and predicting the subsequent frame tokens. This approach unifies our training into a simple task of completing a sequence and yields a powerful joint representation that allows us to reason over the timescale of entire task sequences. The model has the potential to generalize to other tasks if their input/output and goals can be specified by the same tokenized representation. 

\subsection{Data Collection}

Training a self-supervised transformer model requires a large amount of trajectory data. In this work, we use off-policy data to bootstrap our model but preliminary tests show that collecting the on-policy data can reduce sample complexity even further and speed up convergence, at the expense of running a simulator.

Collecting suitable data is not trivial, as our agent needs to see demonstrations of complex behavior. For example, the ObjectNav tasks consists of several implicit sub-tasks, namely exploring the environment, identifying an object of the goal category, and walking towards it.

One solution for this is to collect expert human demonstrations, which inherently contain such complex behavior. We utilize this type of data for the Habitat~\cite{habitat} environment provided by \cite{HabitatWeb}.

Another solution is to collect demonstrations using a classical planner with access to full scene information. Such omniscient experts do not inherently produce high quality trajectories as they know exactly how to get to the target and thus exhibit no exploratory behavior. We utilize this approach for collecting trajectories in the AI2-THOR~\cite{ai2thor} environment as human demonstrations are not available.

\subsection{Tokenization}
\label{sec:token}
To fully exploit the ``predict the next token" training recipe we must tokenize our trajectory data into seqeunces of discrete integers. In NLP this is trivial, as each word can intuitively map to an integer (although in practice more complicated schemes are used). In computer vision, images are often quantized by discrete variational autoencoders, which encode an image as a sequences of several integers, and are capable of reconstructing the image from the sequence. Because of this, they act as slightly lossy reversible transforms, and they allow us to model a proxy for the distribution of action-conditioned future observations, by modeling the distribution over a sequence of just a few future observation tokens.
We utilize the following tokenization schemes to convert our embodied  trajectories into sequences of tokens:

\noindent
\textbf{Image Tokenization} is performed by an off-the-shelf discrete variational autoencoder, namely the VQ-GAN model \cite{vqgan}. While tokenizers that quantize images using 32x32 tokens produce more faithful reconstructions, we choose to use a variety that produces only 16x16 tokens for each image, drastically reducing the computation requirements per frame, allowing us to train with longer sequences and larger models. For our experiments we use the encoder with a vocabulary size of 16,384 but in practice we only saw a marginal visual reconstruction benefits over using a vocabulary size of 1024.

\noindent
\textbf{Pose Tokenization} is performed by discretizing position in the x-axis, position in the y-axis, and rotation into 3 pose tokens. The absolute x and y positions of the agent are binned to the nearest 0.1m box in the range of -100m to +100m for a total vocabulary size of 2000. The absolute rotation of the agent is binned to the nearest 3.6 degrees in the range of 0 to 356.4. Using a relative positioning scheme teaches the model to compute positional offsets from the initial frame instead of some absolute origin which might be unknown in an evaluation environment.

\noindent
\textbf{Action Tokenization} is performed by mapping the following actions to integers: \textit{MoveAhead}, \textit{MoveBack}, \textit{RotateLeft}, \textit{RotateRight}, \textit{LookUp}, and \textit{LookDown} and \textit{Stop} for a total vocabulary size of 7. The translation step is 0.25m and the rotation step is 30 degrees. The action set is the same between the AI2-THOR~\cite{ai2thor} and Habitat~\cite{habitat} environments used for the experiments, while the physical agent dimensions and camera parameters differ.

\noindent
\textbf{Goal Tokenization} is performed by encoding either the relative offset of a target location (for point-goal navigation task) using two positional tokens or an object category (for object-goal navigation) using a single target category token.

\subsection{Architecture} 
We propose a model architecture designed to address two crucial challenges of applying our approach to Embodied AI: long sequences and the input constraints of the various prediction heads.
\newline

\noindent
\textbf{Spatio-Temporal Backbone.}
Embodied tasks contain sequences of several hundred frames. Each one of our frames contains 256 image tokens, 3 pose tokens, and 1 action token, therefore it is not computationally practical to concatenate them into a single long sequence. Instead we attend across space (by which we mean across tokens within one frame) and across time (by which we mean across tokens that occupy the same position in different frames) in alternating layers as in \cite{Bertasius2021IsSA}, see Figure~\ref{fig:model} for details. This mechanism allows us to keep our sequence lengths low while still aggregating information across all the tokens in a trajectory. We achieve this by simply reshaping the output after every layer. We utilize a GPT-style encoder-only transformer model with 8 spatial and 7 temporal layers and an embedding dimension of 512.

\noindent
\textbf{Separate Decoder Heads.}
The second constraint of our problem space is that the decoders can not share all information. For instance, the action head is predicting the action the agent should take next to achieve some goal, while the frame prediction head needs to take this very same action as an input to be able to predict the next frame. Because of this dependency structure, we must take care to not leak information from the future. To that end we apply a diagonal causal mask over the spatio-temporal encoder and decode the pose, action and future frame predictions using three separate decoders for each frame. The decoders are 4 layer transformers with an embedding dimension of 512 and no masking. They each take a different set of tokens as inputs and produce their corresponding predictions as illustrated in Figure~\ref{fig:model}. This architecture allow us to share the vast majority of the parameters across all tasks, as all the decoders utilize the embeddings produced by the spatio-temporal encoder and pass their gradients through it during training.

\noindent
\textbf{CLIP Embeddings.}
Our empirical experimentation has shown that the tokens produced by VQ-GAN are not easily separable by semantic concepts such as the categories of objects the image contains. Because of this, image tokens are not ideal inputs for our model, as they do not give us the full representational power of traditional image encoders, despite being trained on large datasets. To overcome this, we feed our model CLIP embeddings instead of image tokens and simply predict the VQ-GAN tokens of the next frame. A similar strategy is employed by BEiT \cite{beitv2} to learn a strong image encoder.

\subsection{Loss Function}
We optimize our model to jointly predict the pose of the agent, the action it should take towards achieving a specified goal, and the image observation that the agent will obtain if it takes a specified action. We hypothesize that future frame prediction will train our model to synthesize information over a long sequence, since it relies on remembering the visual characteristics of parts the room that our agent has seen a long time ago, as well as understand how specific actions influence the observations that the agent will receive (e.g., moving forward makes objects ahead larger). The pose prediction will help orient our agent and measure the space it moves through. The action prediction produces successful navigation policies for moving across the environment, which is the main goal of our work. Our loss has the following three components, each of which is calculated as the cross entropy between the predicted distribution and the target tokens.

\noindent
\textbf{Frame Loss} is applied to the future frame prediction head, which takes as input the embedding of the current frame, the current pose, and the action our agent will take next and outputs the distribution over the vector-quantized image tokens of the next frame.

\noindent
\textbf{Pose Loss} is applied to the pose prediction head, which takes as input the embedding of the current frame and outputs the distribution over the pose tokens of the agent at this time step.

\noindent
\textbf{Action Loss} is applied to the action prediction head, which takes as input the embeddings of the current frame, the current pose, and a goal and outputs the distribution over actions that the agent should take to achieve the specified goal.

 Initial experiments suggested that learning future frame prediction was most important for developing strong trajectory representations, but we chose to model the pose and actions of the agent as well to unify all three tasks into a single sequence prediction problem.

\subsection{End Tasks}
 We describe our end tasks more concretely, below. In addition to being good benchmarks of progress, these tasks can be combined to achieve complex, multi-stage behavior. 

\noindent
\textbf{Localization:}  Given a sequence of past frames, agent poses and actions as well as the current frame, determine the position and orientation of the agent.

\noindent
\textbf{Future Frame Prediction:} Given a sequence of frames, agent poses and actions, predict the next frame conditioned on the next action. 

\noindent
\textbf{Navigation:} Given a target positional offset from the initial location or a target object as well as all past observations and actions, navigate an agent across a building environment to the target. If the target is a position (PointNav), the task is successful if the agent navigates to the target point and selects the stop action within 0.2m of the specified position. If the target is an object category (ObjectNav), the agent navigates to any instance of the target category and selects the stop action within 1.5m of the object.

\subsection{Data Augmentation}

\noindent
\textbf{Future Frame Masking.}
One interesting detail we encountered was the importance of masking out the entire future frame during training. We attempted to utilize a fully diagonal masking strategy where each token could attend to all the tokens to the left, and so consequentially most tokens of a frame $I_{n+1}$ could attend to all the tokens of $I_n$ as well as some of the tokens of $I_{n+1}$. In the limited experiments we attempted to train our model with this masking strategy, it learned to reconstruct $I_{n+1}$ from the partial information about the frame given to it, instead of learning how taking the action $a_n$ transforms frame $I_n$ into frame $I_{n+1}$. 

\noindent
\textbf{Hindsight Trajectory Superposition.}
There is a major imbalance between the frequency of the `stop' action (when the agent reaches the goal) and all the other actions, as the agent stops only once in each episode. To ensure that our model learns to recognize goal conditions, we modify our action labels such that every state, which is a valid goal condition for any ObjectNav category, is marked as `stop'. We also select a number of states roughly equivalent to the number of ObjectNav goal states and assign them as PointNav goal states, and their corresponding action labels are set to stop. In addition, we modify the goal tokens of all other states to line up with the goal of the next stop state. We can achieve this type of hindsight trajectory modification, because we only modify the goal tokens and action labels, neither of which affect the state of the spatio-temporal encoder. This allows us to artificially superimpose several tasks over a single trajectory during training so we call it \textit{Hindsight Trajectory Superposition (HTP)}, (inspired by Hindsight Experience Replay \cite{Andrychowicz2017HindsightER} which shares conceptual similarities).

%% file: sections/04_experiments.tex
\section{Experiments}

We evaluate ENTL on four tasks: PointNav, ObjectNav, localization, and future frame prediction. We compare our method to some strong, popular baselines. We also provide results of ablation experiments that profile the importance of the future frame prediction loss and show scaling laws. Note that our method only uses RGB images (no depth image or motion sensors).

\subsection{Data Collection}

\noindent \textbf{Environments.} We train and evaluate our approach in two interactive embodied AI simulators: AI2-THOR~\cite{ai2thor} and Habitat~\cite{habitat}. AI2-THOR simulates houses populated with digital assets that model objects and furniture in a house. Habitat utilizes indoor 3D scans of real world buildings to simulate living and working environments. Each simulator has different visual and behavioral characteristics, and we show that our method can be applied to both. Across both, our agent can move forward in 0.25m increments, rotate left or right in place by 30 degree increments, tilt its camera up or down by 30 degree increments and signal when it thinks the task has been complete with a `stop' action.

In the Habitat environment, we train our model on 8M frames collected by human annotators as part of Habitat-Web data~\cite{HabitatWeb}. These are high-quality trajectories because they capture the human intuition involved in exploring an environment, identifying an object as an instance of the goal category, and navigating towards it.

In the AI2-THOR environment, we do not have access to a large dataset of human-collected trajectories, but we can obtain expert trajectories from a large collection of procedurally generated scenes by using ProcTHOR~\cite{ProcTHOR} and an omniscient expert planner. While these trajectories do not contain the exploratory behavior that humans exhibit when interacting with a new environment, they do contain a greater diversity of floorplans that the agent navigates across. We collected a dataset of 20M ProcTHOR frames, consisting of trajectories of no less than 100 steps (frames) each, in which the agent visits every target object in each of 10k train scenes in random order.

\subsection{Training}
We train a model on ProcTHOR data, and a separate model on Habitat data for efficiency (combining the data from these environments did not improve the results significantly). The models have the ENTL architecture with 8 spatial and 7 temporal layers in the encoder and 4 layers in each of three decoders. Each layer has 8 heads, 512 embedding dimensions and a MLP factor of 4. The entire ENTL model has 96M parameters. We train the model at half precision with a batch size of 8 for 500K steps using the AdamW optimizer \cite{Loshchilov2019DecoupledWD} and a learning rate of $1e-4$.

\subsection{Navigation}
We evaluate the performance of our model on the PointNav~\cite{Anderson2018OnEO} and ObjectNav~\cite{batra2020objectnav} tasks. 

\noindent \textbf{PointNav.} We train a version of ENTL on just 1M frames for the PointNav~\cite{Anderson2018OnEO} task. Remarkably, we find that our method is able to achieve performance on par with a PPO baseline method that used 50x more data (Table~\ref{table:pointnav}). Additionally, the baseline uses depth as input, which is known to work better for PointNav, while our model uses RGB only. Our model also has the benefit of producing auxiliary outputs which can be used for future frame position and localization, even when trained on such a small dataset. This shows that ENTL is capable of constructing world models and learning to solve simple tasks using a very small amount of samples. Since the encoder of the PPO baseline is not based on CLIP, we also test a version of ENTL without the CLIP encoder using the VQ-GAN tokens as inputs and targets instead (referred to as ENTL RGB) and find that it does not make a large difference. Finally, we evaluate a CLIP+LSTM imitation learning baseline (IL RGB+CLIP) trained on the same data as ENTL and show that it works significantly worse than our method. 

\begin{table}[h!]
\centering
\small
\tabcolsep=0.03cm
\begin{tabular}{lcccc}
\toprule
Model & Environment & \emph{\#} Frames & Success & SPL \\ 
\midrule
 \textbf{DD-PPO Depth}~\cite{Weihs2020AllenActAF} & RoboTHOR & 50M & 0.93 & 0.88 \\ 
\textbf{IL RGB-CLIP}  & RoboTHOR & 1M & 0.61 & 0.52 \\ 
\textbf{ENTL RGB} (Ours) & RoboTHOR & 1M & 0.92 & 0.87 \\
 \textbf{ENTL RGB+CLIP} (Ours) & RoboTHOR & 1M & 0.95 & 0.91 \\
\bottomrule
\end{tabular}
\caption{\textbf{Point-goal Navigation Results.} We use AllenAct~\cite{Weihs2020AllenActAF} implementation of DD-PPO~\cite{wijmans2019dd} as the baseline.}
\vspace{-0.2cm}
\label{table:pointnav} 
\end{table}

\noindent \textbf{ObjectNav.} The model trained on ProcTHOR data is evaluated on RoboTHOR~\cite{RoboTHOR} val scenes, showcasing the ability of our model to adapt to visual domains outside of the training data. Our approach achieves competitive task success compared to EmbCLIP~\cite{embodied-clip} trained on ProcTHOR~\cite{ProcTHOR} data, which is a state-of-the-art model trained with an order of magnitude more data (typically EmbCLIP success rates has 2-3\% variance). Table~\ref{table:objectnav} shows the results. 

We also evaluate a version of our model trained on the Habitat MP3D Val split consisting of unseen scenes. The model achieves strong performance when trained on 2.5x less data than the imitation learning baseline using Habitat-Web~\cite{HabitatWeb} data (referred to as `IL w/ Hab.-W'). The reason we chose this baseline is that it uses the same type of data as our model. Note that the architecture is identical for both navigation tasks, but we train the model for each task separately. We chose this subset of data since it allowed us to achieve a similar SPL compared to the baseline, while requiring a much lower sample complexity.

\noindent
\textbf{Sequence Learning Baselines.}
We analyze the performance of two popular sequence learning baselines trained on the same data as ENTL. We implement a vanilla version of Decision Transformer \cite{chen2021decision} with a CLIP encoder (for fair comparison), where we embed our current observations into a single token, then predict future observations and rewards. We add an additional l2 loss to model future observation tokens based on current observations and actions. This model performs similarly to our "no future frame prediction" baseline, showcasing the importance of our specific type of future frame prediction loss. We also implemented a version of Dreamer \cite{Hafner2019DreamTC}. We found that this baseline was mostly successful at nearby targets. See Table~\ref{table:objectnav} for the results. We hypothesize that learning to predict future states in the embedding space is not sufficient to obtain a strong understanding of the environment.

\begin{table}[h!]
\centering
\tabcolsep=0.07cm
\small
\begin{tabular}{lcccc}
\toprule
Model & Environment & \emph{\#} Frames & Success & SPL \\ 
\midrule
 \textbf{EmbCLIP}~\cite{ProcTHOR} & RoboTHOR & 200M & 0.52 & 0.22 \\ 
 \textbf{ENTL} (Ours) & RoboTHOR & 20M & 0.46 & 0.16 \\
 \textbf{Dreamer \cite{Hafner2019DreamTC}} & RoboTHOR & 20M & 0.31 & 0.06 \\
 \textbf{Decision Transformer \cite{chen2021decision} } & RoboTHOR & 20M & 0.40 & 0.11 \\
 \midrule
 \textbf{IL w/ Hab.-W}~\cite{HabitatWeb} & Habitat & 22M & 0.23 & 0.06 \\
 \textbf{ENTL} (Ours) & Habitat & 8M & 0.17 & 0.05 \\
\bottomrule
\end{tabular}
 \caption{\textbf{Object-goal Navigation Results.} EmbCLIP~\cite{embodied-clip} is trained on trajctories from ProcTHOR~\cite{ProcTHOR}. IL w/Hab.-W is the imitation learning model from \cite{HabitatWeb} that, similar to our model, uses only RGB images.}
 \vspace{-0.3cm}
  \label{table:objectnav} 
\end{table}
\begin{figure*}[tp]
    \vspace{-0.2cm}
    \centering
    \includegraphics[width=39pc]{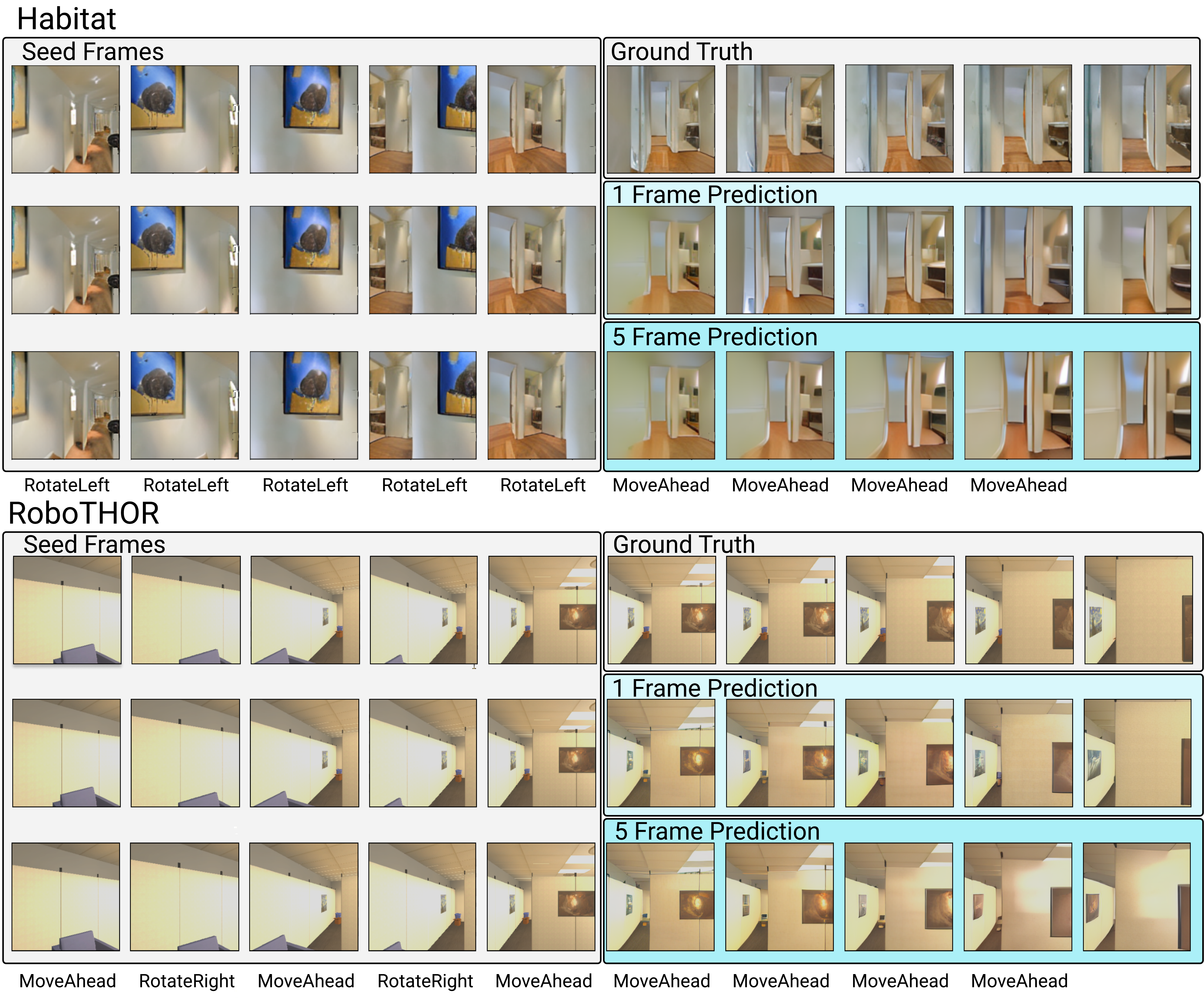}
    \caption{\textbf{Predictions of future frames in unseen scenes.} For each rollout 5 seed frames are given to the model, then it predicts the next 5 steps with and without re-seeding after each step, based on the provided actions. Additional examples in appendix \ref{fig:supp-qualitative}. }
    \label{fig:qualitative}
    \vspace{-0.4cm}
\end{figure*}

\subsection{Localization}
An ENTL agent receives a tokenized representation of its current and past pose as input, and predicts the pose in the next state. Because of this it naturally performs the task of localization. We evaluate the ability of our agent to effectively localize over a long trajectory by predicting the pose tokens for a sequence of 50 steps. We observe a final mean localization error of 0.43m and a std of 0.59m on a dataset of 10,000 RoboTHOR trajectories and an error of 1.12m and std of 1.52m on a dataset of 10,000 Habitat trajectories. A naive baseline that simply estimates the agent pose based on the actions achieves errors of 0.93m and 2.34m on RoboTHOR and Habitat, respectively. The high standard deviations suggest that for some trajectories our model performs very well and for a few it has a high error. Note that movements in these environments are stochastic, and our model does not use any depth or motion sensor.

\subsection{Future Frame Prediction}
\vspace{-0.2cm}
While future frame prediction is not the main focus of our work, the nature of our loss function optimizes this objective as well. To test the quality of future frame prediction, we compute the SSIM score~\cite{wang2004image} of future frame predictions on unseen scenes, and compare it to a naive baseline of simply repeating the last frame. We achieve a SSIM score of 0.361 on 10k Habitat trajectories (compared to the baseline of 0.219) and 0.578 on 10k RoboTHOR trajectories (compared to the baseline of 0.396). Figure~\ref{fig:qualitative} shows generation of 1 and 5 frames into the future. For instance, in the Habitat example, the model learns to predict larger objects as the agent moves forward.

\subsection{Ablations}

\noindent
\textbf{Impact of Various Objectives}
To profile the contribution of the individual objectives during pre-training, we test a version of the ENTL model where the frame loss was set to 0. Without this loss component the model performance drops significantly, highlighting the importance of future frame prediction to the extraction of useful sequence representations. We also profile the effect of omitting the pose prediction loss and find it has negligible impact on performance. Finally we profile the impact of omitting HST label modification and find it has a significant impact. We run these experiments using the smaller 10M model. Refer to Table~\ref{tab:loss-ablations} for the results.

\begin{table}[h!]
\vspace{-0.2cm}
  \centering
  \small
\begin{tabular}{cccc}
\toprule
Model & \emph{\#} Frames & Success & SPL \\
\midrule
 \textbf{w/o Frame Pred. Loss} & 20M & 0.21 & 0.03 \\
 \textbf{w/o HST} & 20M & 0.28 & 0.04 \\
 \textbf{w/o Pose Pred. Loss} & 20M & 0.34 & 0.06 \\
 \textbf{w/ Frame Pred. Loss} & 20M & 0.35 & 0.06 \\
\bottomrule
\end{tabular}
  \caption{\textbf{Frame Prediction Loss.} }
    \vspace{-0.4cm}
  \label{tab:loss-ablations} 
\end{table}

\noindent
\textbf{Scaling Laws.}
We analyze the effects of model size on final performance. We find that increasing model size monotonically improves performance on the RoboTHOR ObjectNav task. We train a version of our model using 10M and 50M parameters, respectively, using the same dataset and training schedule. Refer to Table~\ref{tab:scaling-laws-ablations} for the results.


\begin{table}[h!]
  \centering
  \small
\vspace{-0.2cm}
\begin{tabular}{ccc}
\toprule
Model & \emph{\#} Success & SPL \\
\midrule
 \textbf{ENTL-10M} & 0.35 & 0.06 \\
 \textbf{ENTL-50M} & 0.42 & 0.13 \\
 \textbf{ENTL-100M} & 0.46 & 0.16 \\
\bottomrule
\end{tabular}
  \caption{\textbf{Scaling Laws.} }
  \vspace{-0.5cm}
  \label{tab:scaling-laws-ablations} 
\end{table}

%% file: sections/05_conclusion.tex
\section{Conclusion}
\vspace{-0.2cm}
We introduced ENTL: Embodied Navigation Trajectory Learner, a new approach for sequence representation learning for embodied navigation tasks. We presented a strategy for converting embodied trajectories into sequences of discrete tokens, and demonstrated that it is an effective approach in practice. We present a model architecture that effectively synthesized information across both space and time, and trained various instance of that model on canonical embodied navigation tasks. We compare our approach to baseline methods to demonstrate its advantages. Our approach demonstrated significant gains in sample complexity, as well as producing useful auxiliary outputs. For example, our model is able to visualize future frames which can be useful for explainability or planning, and localize the agent. Our architecture does not make assumptions about the specific tasks we studied, so it is generic and can be used for other embodied tasks.

%% file: sections/06_supp.tex
\newpage
\section*{Appendix}
\section{Results for Different Image Encoders}
To evaluate whether our model provides improvements over other visual backbones, we train a version of ENTL that uses the Masked Vision Pre-training (MVP) \cite{Xiao2022} encoder pre-trained on embodied robotics data. We evaluate the model on the ObjectNav task, and as Table~\ref{tab:encoder-ablation} shows the CLIP based model outperforms the MVP based one. 

\begin{table}[h!]
\centering
\small
\begin{tabular}{cccc}
\toprule
 Model & \emph{\#} Frames & Success & SPL \\
 \midrule
 \textbf{ENTL w/ MVP}~\cite{Xiao2022} & 20M & 0.47 & 0.15 \\ 
 \textbf{ENTL w/ CLIP}~\cite{embodied-clip} & 20M & 0.50 & 0.18 \\
\bottomrule
\end{tabular}
 \caption{\textbf{Encoder Ablation.} }
  \label{tab:encoder-ablation} 
\end{table}

\section{ProcTHOR Data Collection}
We collect 20 Million frames of demonstrations in the ProcTHOR environment which we use for training the joint ObejctNav and PointNav ETL model. Since there is no large scale human demonstration dataset in this environment, we collect trajectories guided by a planner with full information about the scene. The planner plots the optimal path from the current location of the agent to a valid goal state for a given object category. This planner is implemented in the Unity environment upon which AI2-THOR is built. A goal state is defined as any state in which the agent is within 1.5m of the target object, and the object is visible (as defined by the RoboTHOR ObjectNav challenge).

Since a direct trajectory from the current location to a goal does not capture any exploratory behavior, we instead structure our trajectories as tours of every target object in a given scene. Each trajectory is obtained by navigating the agent from a random starting location to a goal object, then from that object to every other goal object in the current scene in random order. This way the agent walks around an entire room in every trajectory.

At each step we record the RGB frame, the action that was taken, the agent pose and the list of all ObjectNav target objects that are currently in range of the agent. Only RoboTHOR ObjectNav challenge goal objects are considered as goal objects. 

Any state can also serve as a PointNav goal state, so we do not collect PointNav trajectories explicitly.

\section{Additional Frame Prediction Examples}
\begin{figure*}[tp]
    \centering
    \caption{Additional predictions of future frames in unseen RoboTHOR scenes. For each rollout 3 seed frames are given to the model, then it predicts the next 5 steps with and without re-seeding after each step, based on the provided actions. }
    \vspace{1.0cm}
    \label{fig:supp-qualitative}
    \includegraphics[width=39pc]{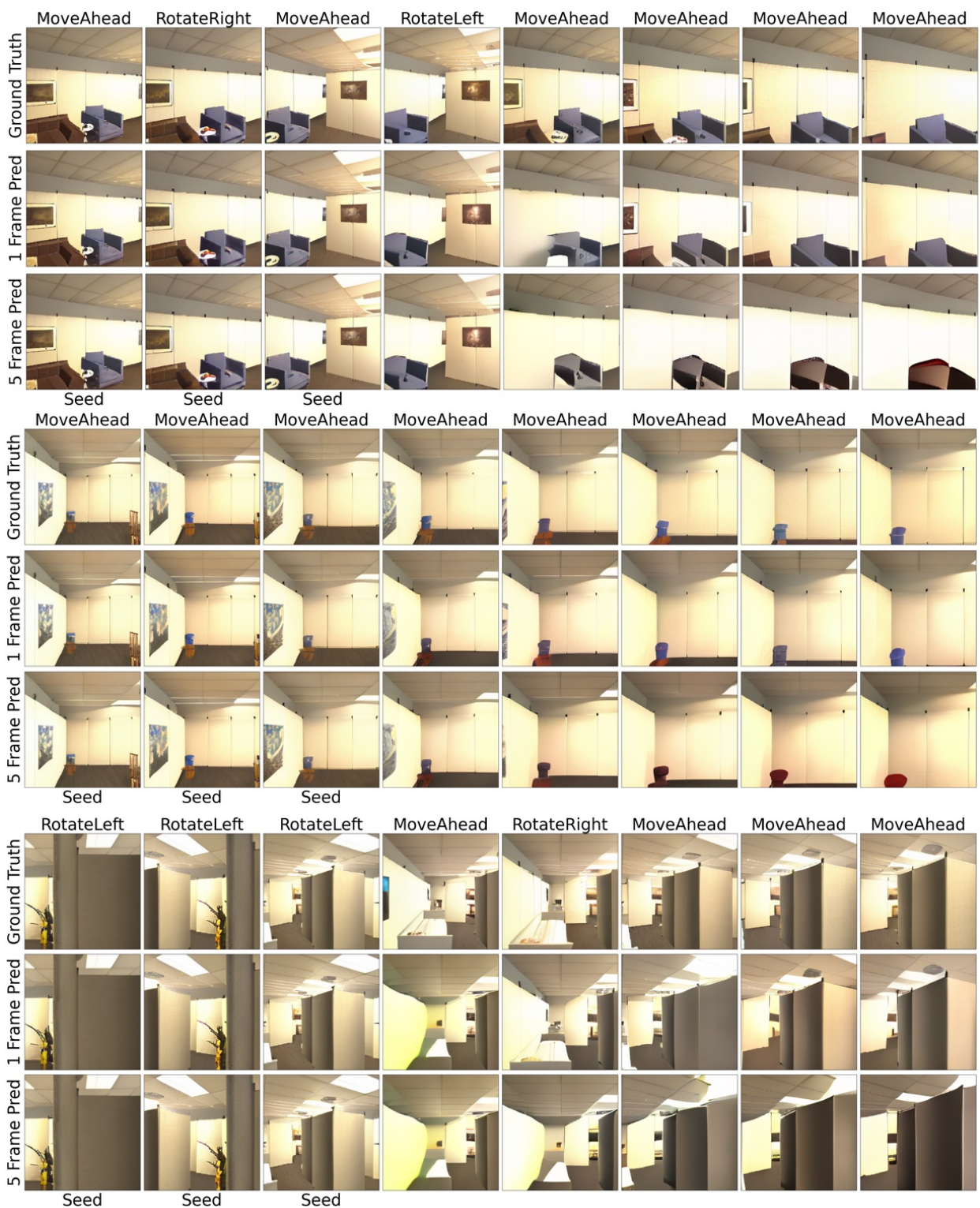}
\end{figure*}

Additional rollouts produced by the ENTL model. As shown the model is able to predict fairly well if the room is empty or objects are given in the seed frames, but performs poorly when hallucinating new objects. Additionally, each consecutive predicted frame suffers from an accumulation of errors as the predicted image tokens have to be decoded into an RGB image by VQ-GAN, then passed through CLIP to predict the next frame. The results are shown in \ref{fig:supp-qualitative}.

\begin{figure*}[tp]
    \centering
    \includegraphics[width=39pc]{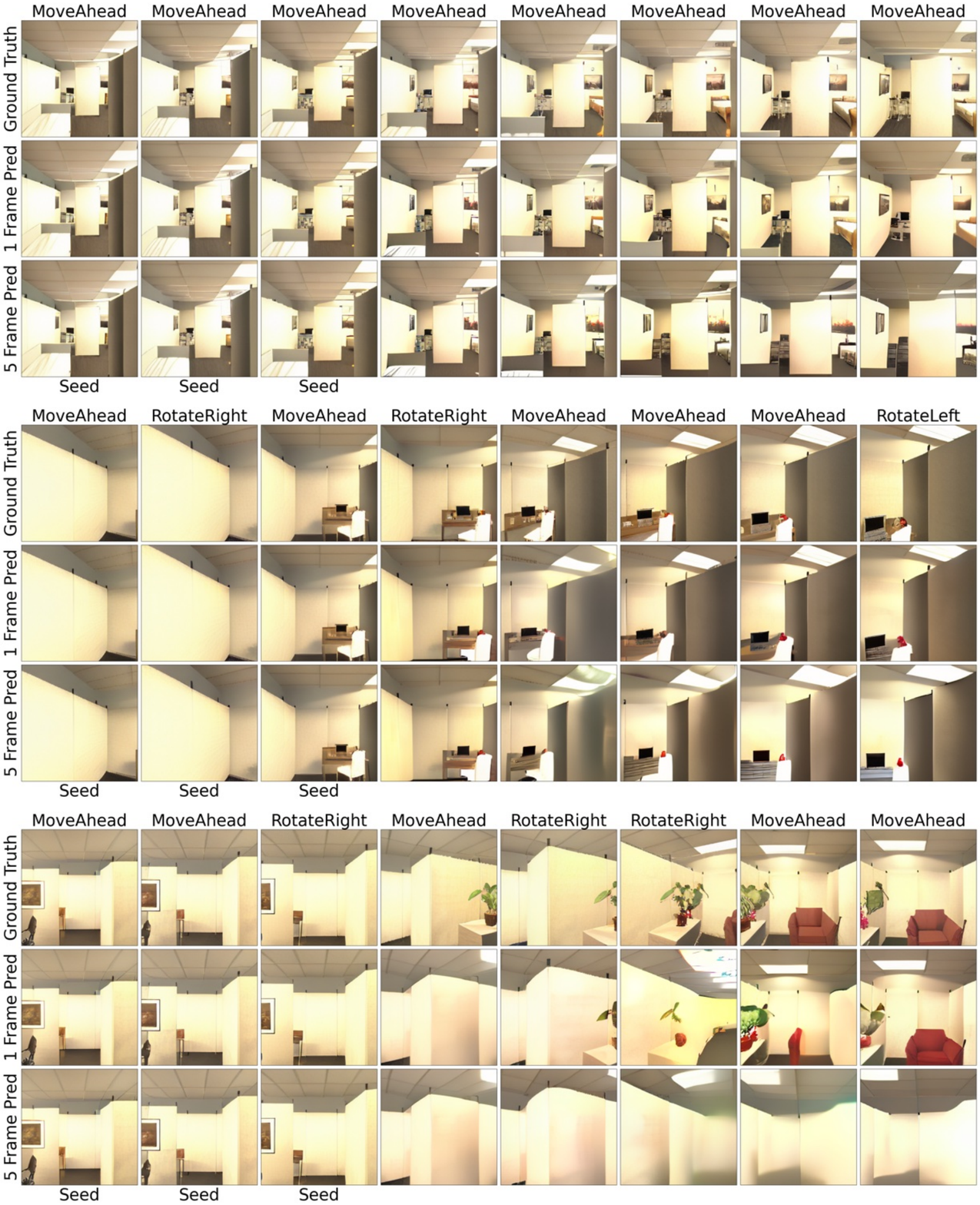}
\end{figure*}